\documentclass[]{interact}

\usepackage{epstopdf}
\usepackage{subfigure}
\usepackage{booktabs}

\usepackage{natbib}
\bibpunct[, ]{(}{)}{;}{a}{}{,}
\renewcommand\bibfont{\fontsize{10}{12}\selectfont}

\usepackage{graphicx}
\usepackage{amsmath,amssymb,amsfonts}
\usepackage{amsthm}
\usepackage{mathrsfs}
\usepackage{xcolor}
\usepackage{algorithm}
\usepackage{algorithmicx}
\usepackage{algpseudocode}

\theoremstyle{plain}

\theoremstyle{definition}

\theoremstyle{remark}

\begin{document}


\title{A Transfer Learning-Based Method for Water Body Segmentation in Remote Sensing Imagery: A Case Study of the Zhada Tulin Area}

\author{
	\name{Haonan Chen\textsuperscript{a} and Xin Tong\textsuperscript{b}\thanks{CONTACT Xin Tong. Email: tongxin@mail.nwpu.edu.cn}}
	\affil{\textsuperscript{a}Engineering College, Tibet University, Lhasa, 850000, China; \textsuperscript{b}School of Physical Science and Technology, Northwestern Polytechnical University, Xi'an, 710072, China}
}

\maketitle

\begin{abstract}
The Tibetan Plateau, known as the 'Asian Water Tower', faces significant water security challenges due to its high sensitivity to climate change. Advancing Earth observation for sustainable water monitoring is thus essential for building climate resilience in this region. This study proposes a two-stage transfer learning strategy using the SegFormer model to overcome domain shift and data scarcity—key barriers in developing robust AI for climate-sensitive applications. After pre-training on a diverse source domain, our model was fine-tuned for the arid Zhada Tulin area. Experimental results show a substantial performance boost: the Intersection over Union (IoU) for water body segmentation surged from 25.50\% (direct transfer) to 64.84\%. This AI-driven accuracy is crucial for disaster risk reduction, particularly in monitoring flash flood-prone systems. More importantly, the high-precision map reveals a highly concentrated spatial distribution of water, with over 80\% of the water area confined to less than 20\% of the river channel length. This quantitative finding provides crucial evidence for understanding hydrological processes and designing targeted water management and climate adaptation strategies. Our work thus demonstrates an effective technical solution for monitoring arid plateau regions and contributes to advancing AI-powered Earth observation for disaster preparedness in critical transboundary river headwaters.
\end{abstract}


\section*{Funding}
The authors declare that no funds, grants, or other support were received during the preparation of this manuscript.

\section{Introduction}\label{sec1}

Surface water bodies are crucial natural elements, and accurately mapping their spatial distribution is vital for sustainable water resource management, climate resilience building, and disaster risk assessment \citep{mcfeeters1996use}. High-resolution remote sensing imagery has become a primary data source for this task.The integration of artificial intelligence with Earth observation systems represents a significant advancement in our capacity to support climate resilience and disaster risk reduction efforts. Traditional methods like the Normalized Difference Water Index (NDWI) \citep{mcfeeters1996use, xu2006modification} and threshold segmentation \citep{otsu1975threshold} often struggle in complex environments where water's spectral features are similar to the background, frequently requiring support from other indices like NDVI \citep{rouse1973paper} or SAVI \citep{huete1988soil}.

The advent of deep learning has revolutionised remote sensing image analysis,offering new pathways for AI-driven Earth observation systems that support sustainability goals \citep{goodfellow2016deep, zhu2017deep, ma2019deep, zhang2016deep}. Convolutional Neural Networks (CNNs) \citep{lecun2002gradient}, from early Fully Convolutional Networks (FCNs) \citep{long2015learning} to advanced architectures like U-Net \citep{ronneberger2015u}, SegNet \citep{badrinarayanan2017segnet}, and DeepLab \citep{chen2017deeplab, chen2018encoder}, have significantly improved segmentation accuracy. More recently, Transformer-based models such as the Swin Transformer \citep{liu2021swin} and SegFormer \citep{xie2021segformer}, leveraging the global modelling power of the attention mechanism \citep{vaswani2017attention}, have demonstrated outstanding performance. These methods, along with other innovations like multi-attention networks \citep{li2021multiattention}, non-local networks \citep{wang2018non}, and hybrid Transformer-CNNs \citep{zhang2022transformer}, have greatly enhanced the automatic identification of ground features.

Despite these technological advancements, applying deep learning to real-world remote sensing is challenged by pressing geoscientific needs, particularly in the context of climate change. The Tibetan Plateau, for instance, is warming at approximately twice the global average rate \citep{you2011changes, chen2015qing}, triggering significant hydrological changes like accelerated glacier melt and altered river runoff.These changes increase the frequency of extreme weather events, including flash floods and droughts, making accurate water body monitoring essential for disaster risk reduction and climate adaptation strategies. Traditional sparse monitoring stations are insufficient for capturing these dynamics. High-resolution remote sensing thus offers an indispensable tool for large-scale monitoring, which is crucial for understanding the plateau's evolving water cycle and assessing its sustainability.

This study focuses on a particularly challenging yet critical region: the Zhada Tulin (Zhada Earth Forest) in the Ali Prefecture of Tibet. As a typical arid landscape in the northwestern Tibetan Plateau \citep{ding2006qing}, its water systems are sensitive indicators of regional climate shifts. The area's Xiangquan River is a vital headwater for a transboundary river system, directly impacting water security in downstream South Asia. Moreover, the unique 'earth forest' geomorphology, a product of Quaternary tectonic uplift and climate change, makes it a natural laboratory for studying the plateau's environmental evolution. Given the region's vulnerability to climate-related hazards and its importance for regional water security, developing robust monitoring capabilities is crucial for both scientific understanding and disaster preparedness.

However, the practical application of deep learning models in such specific regions is severely hampered by two core issues. First, these models rely on large, high-quality annotated datasets, which are costly and labour-intensive to produce for remote sensing imagery. Second, a model trained in one geographical area often suffers a sharp performance decline when applied to another due to variations in imaging conditions and landscape features—a phenomenon known as "domain shift" \citep{pan2010q, yosinski2014transferable}.

To address these challenges, transfer learning provides an effective solution. While the 'pre-train and fine-tune' paradigm using ImageNet \citep{krizhevsky2017imagenet} is standard, its direct application to remote sensing yields limited results due to the unique characteristics of satellite imagery \citep{tuia2016domain}. Therefore, designing transfer learning strategies tailored for remote sensing applications remains a key research area.

Against this background, this study proposes a two-stage transfer learning strategy based on the SegFormer model. We first train a robust base model on a diverse source domain and then fine-tune it using a small, targeted dataset from the Zhada Tulin area. This 'general-to-specific' approach aims to achieve high-precision water body segmentation under small-sample and domain-shift conditions. This research not only provides a technical solution for water information extraction in the challenging Zhada region but also contributes to advancing Earth observation capabilities for climate resilience and sustainable water management in climate-sensitive environments.

\section{Study Area and Methods}\label{sec:methods}

To systematically evaluate the effectiveness of a two-stage transfer learning strategy for remote sensing water body segmentation, this study established a data scheme involving a diverse source domain and a unique target domain, along with corresponding procedures for model training, implementation, and evaluation.

\subsection{Study Area and Datasets}\label{subsec:data}

\begin{figure}[t!]
	\centering
	\includegraphics[width=0.8\textwidth]{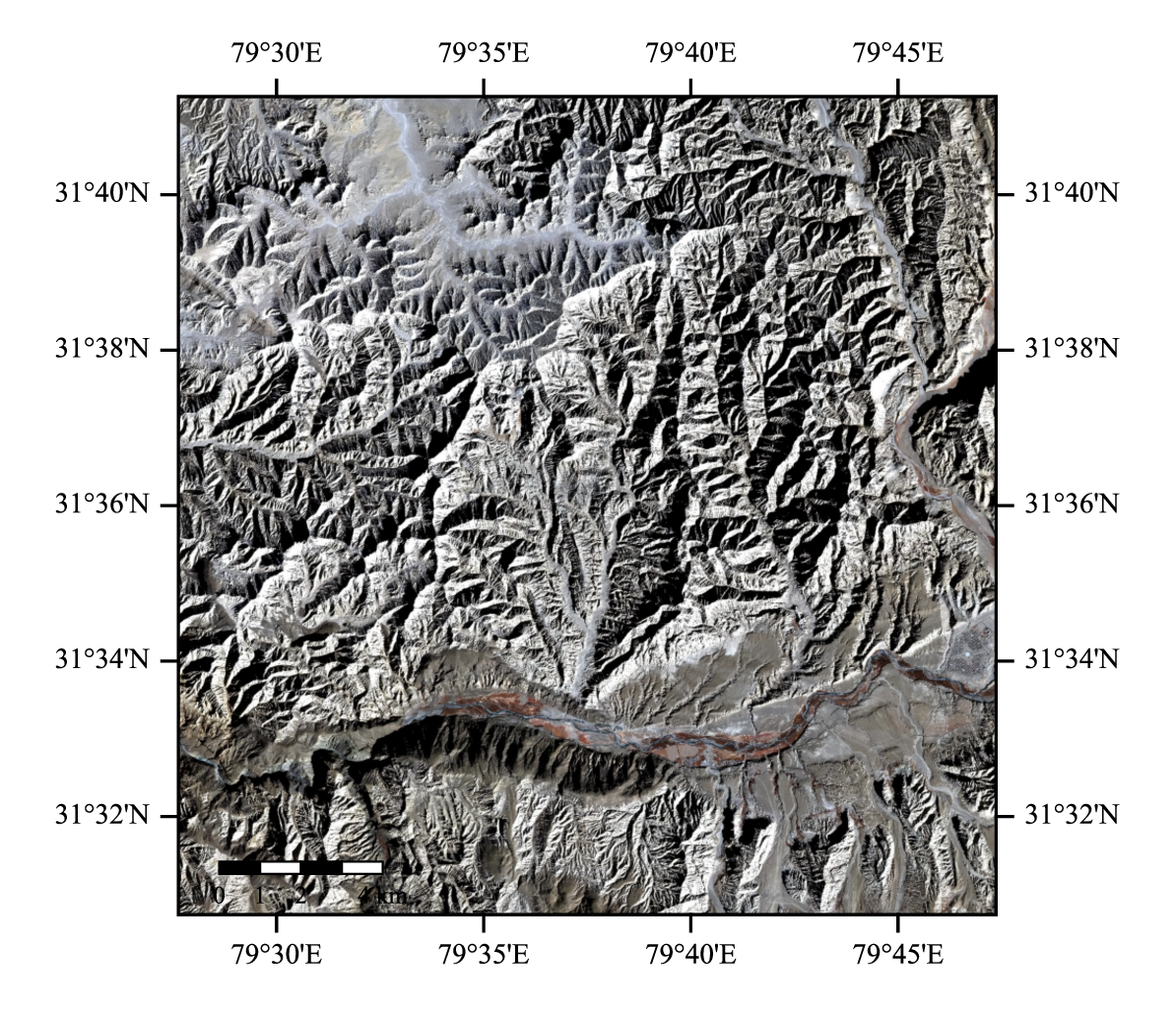} 
	\caption{Geographic location map of the study areas, highlighting the target domain in the Zhada Tulin area.}\label{fig:figure-1}
\end{figure}

The datasets for this study consist of a diverse source domain (Dataset A) designed for robust pre-training and a highly challenging target domain (Dataset B) for evaluating the model's adaptability.

\textbf{Source Domain (Dataset A).} The source domain was constructed to provide rich and generalizable learning samples. The imagery is primarily sourced from China's Gaofen-7 (GF-7) satellite, whose sub-meter resolution (0.8-meter panchromatic and 3.2-meter multispectral) provides an excellent data foundation for fine-grained feature recognition. The spatial coverage is extensive, focusing on diverse plateau and mountainous environments, including arid lakes in Xinjiang, salt lakes in Qinghai, plateau lakes in Tibet, and mountain lakes and rivers in the Rocky Mountains of the United States. This high degree of diversity in geography and water body types ensures that a model trained on this dataset learns universal features of water bodies, possessing strong generalisation potential.

\textbf{Target Domain (Dataset B): The Zhada Tulin} The target domain focuses on a unique and challenging geographical unit---the Zhada Tulin (Zhada Earth Forest) in the Ali Prefecture of Tibet, China (Fig.~\ref{fig:figure-1}). The imagery for this region is sourced from the Gaofen-2 (GF-2) satellite, with a resolution consistent with GF-7, ensuring scale consistency for cross-dataset comparison. The Zhada Tulin is an ideal testbed for our transfer learning strategy due to its distinct geoscientific significance and complex remote sensing characteristics.

\begin{itemize}
    \item \textbf{Hydrological and Geological Features:} Located north of the Bangong-Nujiang Suture Zone, the Zhada area features complex geological structures, primarily composed of Pliocene-Quaternary lacustrine and fluvial sediments \citep{zhu2007xizang}. Situated at an altitude of 3500-4200m, it experiences a high-altitude temperate arid climate with annual precipitation under 200mm and evaporation exceeding 2000mm, making water resources extremely scarce. The water systems mainly consist of the Xiangquan River, its tributaries, seasonal gully runoff, and a few spring outlets \citep{fang2003qing}. Despite their small scale, these water bodies serve critical ecological functions: they create oasis effects that support unique plant communities (e.g., \textit{Tamarix}, \textit{Salix}) and sustain regional biodiversity \citep{zhang2023oceanic}, while also playing a vital role in soil conservation against wind erosion.

	\item \textbf{Climatic Response and Research Value:} Over the past three decades, the hydrology of the Zhada region has undergone significant changes in response to global warming, a trend consistent with broader observations across High Asia \citep{immerzeel2010climate, lutz2014consistent}. These changes manifest as earlier snowmelt-driven runoff, an increased frequency of extreme events like flash floods, and the shrinkage of small lakes. Such dynamics make the Zhada Tulin an ideal case for studying the impact of climate change on arid plateau water cycles. Therefore, precise water body monitoring in this sensitive area is of great scientific value.

    \item \textbf{Challenges for Remote Sensing Segmentation:} The unique environment of the Zhada Tulin presents significant challenges that constitute a domain shift from the source domain. These challenges include: (1) unique morphology, with water systems flowing through narrow, irregular gullies; (2) high turbidity, as rivers carry large amounts of sediment, altering their spectral characteristics; and (3) a complex background, where water bodies are spectrally and geomorphologically coupled with surrounding exposed sediments. These factors make the Zhada Tulin a perfect testbed for validating the effectiveness of the proposed transfer learning strategy.
\end{itemize}

\subsection{Two-Stage Transfer Learning Framework}\label{subsec:framework}

To address the domain shift problem and leverage prior knowledge to enhance segmentation accuracy in the small-sample region, this study designed a two-stage transfer learning framework, as illustrated in Fig.~\ref{fig:framework}. The core idea of this framework is to first train a robust foundational model (Model A1) on a feature-rich source domain dataset, and then use it as a pre-trained model for fine-tuning on the small-sample target domain dataset to obtain a final, high-precision model (Model A2) adapted to the target region. The main steps are as follows:

\begin{figure}[b!]
	\centering
	\includegraphics[width=0.98\textwidth]{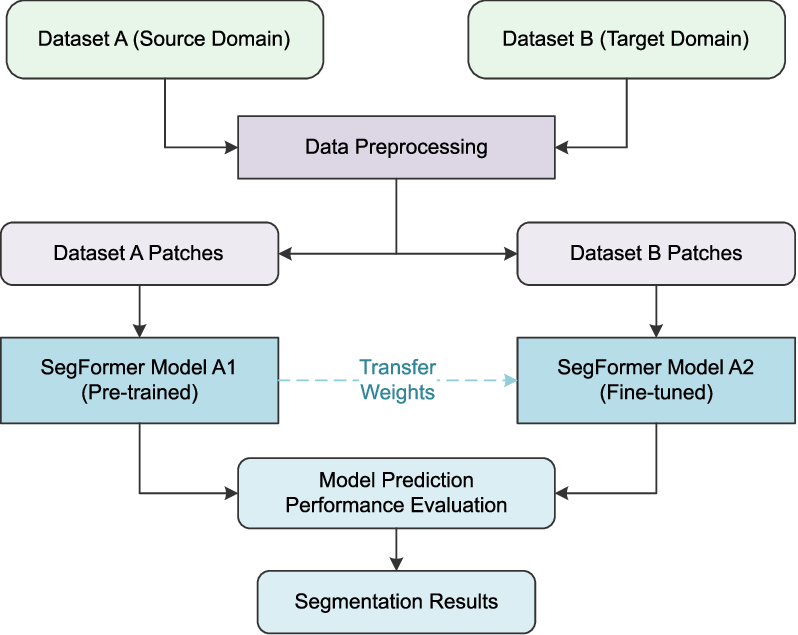} 
	\caption{The overall technical framework of the two-stage transfer learning strategy.}\label{fig:framework}
\end{figure}

\begin{enumerate}
	\item \textbf{Stage 1: Source-Domain Pre-training.} The SegFormer model is thoroughly trained using the source domain dataset (Dataset A). The objective of this stage is to enable the model to learn universal spatial and textural features of water bodies, resulting in a foundational model (Model A1) with strong generalisation capabilities. To accelerate convergence, the encoder component is initialized with ImageNet pre-trained weights.
	
	\item \textbf{Stage 2: Target-Domain Fine-tuning.} All weights of the foundational model A1 are used as initialization parameters for a new model. This model is then secondarily trained (fine-tuned) on the limited-sample target domain dataset (Dataset B). This stage aims to adapt the model to the specific visual characteristics and complex background of the target domain, ultimately producing an optimized segmentation model (Model A2) for the Zhada Tulin area.
	
	\item \textbf{Model Performance Evaluation.} To validate the effectiveness of the proposed strategy, a comprehensive quantitative and qualitative evaluation is conducted on the validation set of the target domain. The performance of the directly transferred model (A1), a baseline model trained from scratch, and the fine-tuned final model (A2) are rigorously compared.
\end{enumerate}

\begin{figure}[htbp]
	\centering
	\includegraphics[width=\textwidth]{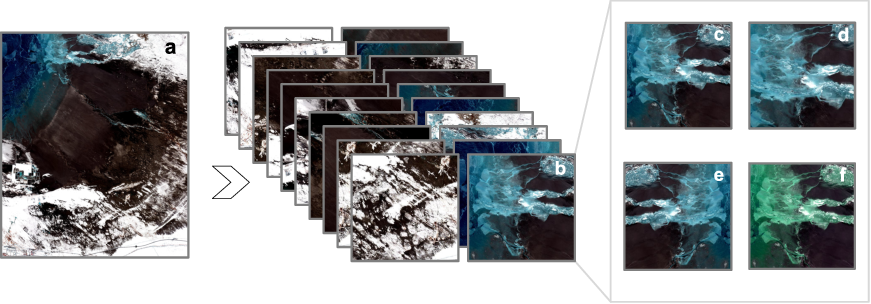} 
	\caption{Diagram of the data preprocessing pipeline: (a) Original remote sensing image; (b) Sliding window cropping into 512$\times$512 patches; (c)-(f) Data augmentation on training samples.}\label{fig:preprocessing}
\end{figure}

To implement this framework, all original remote sensing images must undergo a standardised sample construction process (Fig.~\ref{fig:preprocessing}). This process involves two main steps: sliding window cropping and data augmentation, designed to convert the raw data into patches suitable for model training.

To ensure data quality and simulate real-world challenges, we retained the inherent radiometric differences between the GF-7 and GF-2 raw data products without additional calibration. This approach preserves the natural spectral variations across sensors, enhancing the model's robustness in practical multi-source remote sensing applications where data fusion from different satellites is common.

\textbf{Sliding Window Cropping.} To standardize the input size and effectively augment the number of samples, this study employs an overlapping sliding window method to synchronously crop the original images and their corresponding vector label maps. Specifically, a 512$\times$512 pixel window (Fig.~\ref{fig:preprocessing}b) is moved across the image with a stride of 128 pixels (i.e., a 25\% overlap). The overlap ensures the integrity of features across different samples and avoids the loss of boundary information caused by hard cutting. After cropping, only image patches containing water pixels and their corresponding labels are retained to build an effective training set.

\textbf{Data Augmentation.} To enhance the model's generalisation ability and effectively mitigate overfitting, a series of data augmentation techniques (Fig.~\ref{fig:preprocessing}c-f) are applied exclusively during the training phase. The employed strategies include:
\begin{itemize}
	\item \textit{Geometric transformations:} Random horizontal flipping, and random scaling (with a factor in the range [0.5, 2.0]) followed by a random crop back to 512$\times$512 pixels.
	\item \textit{Photometric distortions:} Random adjustments to the brightness, contrast, and saturation of the image to simulate different lighting and atmospheric conditions.
\end{itemize}
These operations enable the model to learn more robust, deep features of water bodies that are invariant to changes in scale, angle, and illumination.

Through this sample construction pipeline, the final training and validation sets for the source and target domains were obtained. Specifically, 3,875 valid 512$\times$512 pixel patches were cropped from 207 original GF-7 images for the source domain (Dataset A), while 180 patches were obtained from 20 original GF-2 images for the target domain (Dataset B). Subsequently, to ensure an objective and consistent evaluation, both datasets were randomly split into training and validation sets at a 9:1 ratio. It is noteworthy that the training set for the target domain B contains only 162 samples (180 $\times$ 0.9), which further highlights the "small-sample" challenge in this specific region and underscores the necessity of the proposed transfer learning strategy.

\subsection{Semantic Segmentation Model}\label{subsec:model}

\begin{figure*}[htbp]
	\centering
	\includegraphics[width=1.0\textwidth]{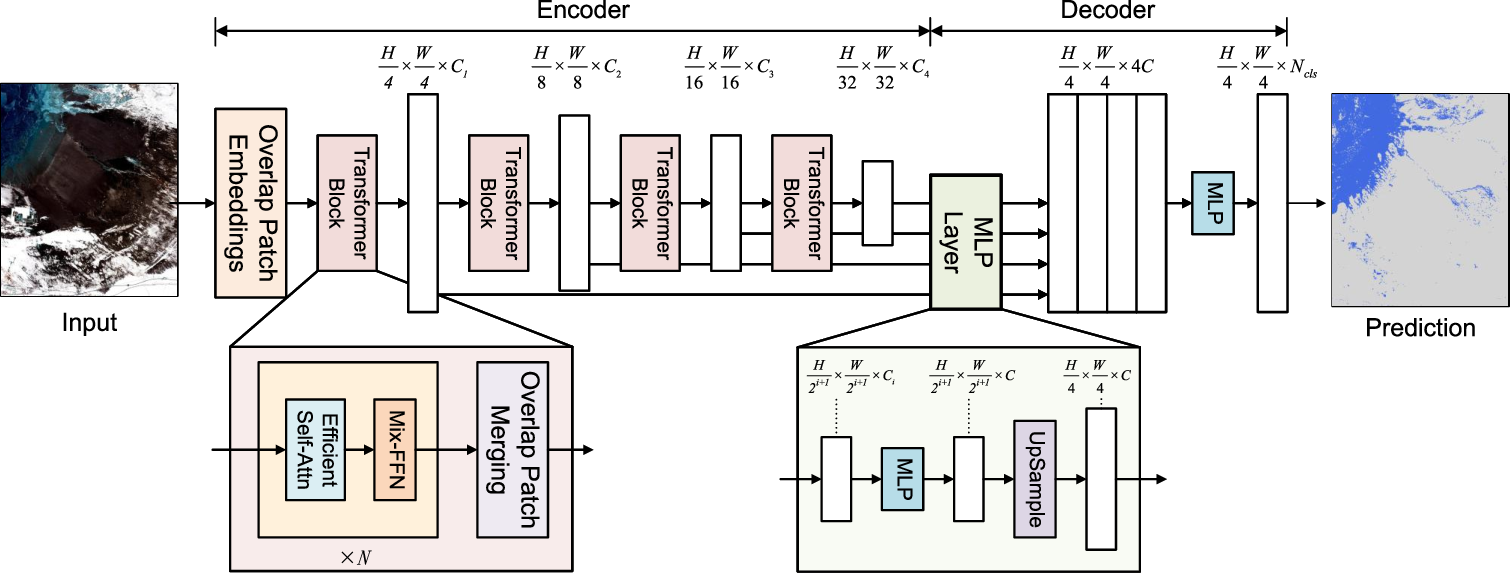} %
	\caption{Architecture of SegFormer, including a hierarchical Transformer encoder and an all-MLP decoder \citep{xie2021segformer}.}\label{fig:segformer}
\end{figure*}

This study selected SegFormer \citep{xie2021segformer} as the core semantic segmentation network for the precise extraction of water bodies from remote sensing imagery. SegFormer is a highly efficient model based on a pure Transformer architecture, striking an excellent balance between accuracy and efficiency, making it particularly suitable for processing high-resolution remote sensing images. The model primarily consists of a hierarchical Transformer encoder and a lightweight all-MLP (Multilayer Perceptron) decoder head, as shown in Fig.~\ref{fig:segformer}.

The encoder of SegFormer employs the Mix Transformer (MiT) \citep{cui2022mixformer} series as its backbone; this study specifically uses the MiT-B5 version. Unlike the traditional Vision Transformer (ViT) \citep{dosovitskiy2020image}, MiT achieves efficient multi-scale feature extraction. It can effectively extract feature representations at different scales from the input image, which is crucial for identifying water bodies of various sizes and shapes. Through an improved self-attention mechanism and a position-encoding-free design, MiT significantly enhances computational efficiency and adaptability to different input sizes. This design paradigm, which has seen successful application in remote sensing image processing \citep{wang2022unetformer}, is well-suited for handling the complex spatial relationships within remote sensing images.

For the decoder, SegFormer innovatively uses a very simple all-MLP head. This decoder head can efficiently aggregate the multi-level features output by the encoder and generate the final pixel-level segmentation map with low computational cost.

By virtue of its powerful multi-scale feature capturing capabilities, excellent computational efficiency, and robust support for high-resolution inputs, SegFormer was chosen as the foundational model for the water body segmentation and subsequent transfer learning exploration in this study.

\subsection{Model Training and Implementation}\label{subsec:training}

The implementation and training of all models in this study were conducted based on the PyTorch \citep{paszke2019pytorch} deep learning framework within a unified hardware environment to ensure the comparability of the results.

To effectively guide the training process of the SegFormer model and address issues such as class imbalance (e.g., significantly more background pixels than water pixels) and ambiguous object boundaries in the water body segmentation task, this study employs a compound loss function. This loss is a linear combination of a weighted Cross-Entropy Loss and a Dice Loss \citep{milletari2016v}.

The Dice Loss originates from the Dice Similarity Coefficient (DSC) and directly measures the overlap between the model's prediction and the ground truth. It is inherently robust to class imbalance and performs well in optimizing segmentation boundaries. The inclusion of Dice Loss encourages the model to generate results with better spatial continuity and clearer boundaries. Its formula is as follows:
\begin{equation}\label{eq:dice_loss}
	\mathcal{L}_{\text{Dice}} = 1 - \frac{2\sum_{i=1}^{N}{p_{i} y_{i}} + \varepsilon}{\sum_{i=1}^{N}{p_{i} + \sum_{i=1}^{N}{y_{i}} + \varepsilon}}
\end{equation}
where $\mathcal{L}_{\text{Dice}}$ is the continuous form of the Dice loss; $p_{i} \in [0,1]$ is the probability predicted by the model for the positive class at pixel $i$; $y_{i} \in \{0,1\}$ is the ground truth label for pixel $i$; $N$ is the total number of pixels; and $\varepsilon$ is a smoothing term to prevent division by zero.

The Cross-Entropy Loss \citep{zhang2018generalized}, a standard pixel-level classification loss for image segmentation, aims to minimize the difference between the predicted pixel class probability distribution and the ground truth. To further mitigate the imbalance between foreground (water) and background pixels, different weights were assigned to each class. Specifically, the weight for the background class was 0.2289, and for the water class, it was 0.7711. By assigning a higher loss weight to the less frequent water class, the model is guided to pay more attention to its correct identification. For our binary semantic segmentation task, the formula is:
\begin{equation}\label{eq:bce_loss}
	\mathcal{L}_{\text{BCE}} = -\frac{1}{N}\sum_{i=1}^{N}{[y_{i}\log(p_{i}) + (1-y_{i})\log(1-p_{i})]}
\end{equation}
where $\mathcal{L}_{\text{BCE}}$ is the binary cross-entropy for our task. This study aims to leverage the advantages of both cross-entropy loss in pixel-wise classification and Dice loss in handling class imbalance and optimizing regional overlap, thereby enhancing the overall segmentation performance.

The core training parameters are configured as follows:
\begin{enumerate}
	\item \textbf{Optimizer:} The AdamW optimizer \citep{loshchilov2017decoupled} was used with an initial learning rate set to $6 \times 10^{-6}$ and a weight decay of 0.01.
	\item \textbf{Learning Rate Schedule:} The training was conducted for a total of 20,000 iterations. A learning rate schedule with a linear warm-up was employed (the learning rate linearly increased from a factor of $1 \times 10^{-6}$ of the initial learning rate to the initial learning rate over the first 1,500 iterations), followed by a polynomial decay until the end of training (with a minimum learning rate of $1 \times 10^{-5}$).
	\item \textbf{Batch Size:} Set to 6.
\end{enumerate}

\subsection{Performance Evaluation Metrics}\label{subsec:metrics}

To comprehensively and objectively evaluate the performance of the different models in segmenting water bodies from remote sensing imagery, a series of standard metrics widely used in the field of semantic segmentation were selected. These metrics measure the consistency between the model's predictions and the ground truth from various dimensions. They are typically calculated based on four fundamental quantities for a binary classification problem (water and background):
\begin{itemize}
	\item \textbf{True Positives (TP):} The number of water pixels correctly predicted as water.
	\item \textbf{False Positives (FP):} The number of background pixels incorrectly predicted as water (misdetections).
	\item \textbf{True Negatives (TN):} The number of background pixels correctly predicted as background.
	\item \textbf{False Negatives (FN):} The number of water pixels incorrectly predicted as background (missed detections).
\end{itemize}

The specific evaluation metrics are defined as follows:
\begin{enumerate}
	\item \textbf{Intersection over Union (IoU):} Measures the overlap between the predicted segmentation area and the ground truth area. It is one of the most commonly used core metrics in semantic segmentation.
	\begin{equation}\label{eq:iou}
		\text{IoU} = \frac{\text{TP}}{\text{TP} + \text{FP} + \text{FN}}
	\end{equation}
	
	\item \textbf{Precision (P):} Represents the proportion of pixels correctly identified as water among all pixels predicted as water (also known as correctness).
	\begin{equation}\label{eq:precision}
		\text{Precision} = \frac{\text{TP}}{\text{TP} + \text{FP}}
	\end{equation}
	
	\item \textbf{Recall (R):} Represents the proportion of actual water pixels that were correctly identified by the model (also known as completeness).
	\begin{equation}\label{eq:recall}
		\text{Recall} = \frac{\text{TP}}{\text{TP} + \text{FN}}
	\end{equation}
	
	\item \textbf{F1-score:} The harmonic mean of Precision and Recall, used to measure the overlap between the prediction and the ground truth.
	\begin{equation}\label{eq:f1}
		\text{F1-score} = 2 \times \frac{\text{Precision} \cdot \text{Recall}}{\text{Precision} + \text{Recall}}
	\end{equation}
\end{enumerate}

\section{Results}\label{sec:results}

\subsection{Source Domain Model Performance}\label{subsec:source_perf}

To ensure the effectiveness of the subsequent transfer learning, the performance of the foundational model A1, trained on the source domain dataset A, was first evaluated. The segmentation performance of Model A1 on the validation set of the source domain is shown in Table~\ref{tab:source_perf}.

\begin{table}[h!]
	\tbl{Performance of the source domain model A1 on its validation set.}
	{\begin{tabular}{@{}lcccc@{}}
			\toprule
			\textbf{Class}      & \textbf{IoU (\%)} & \textbf{F1-score (\%)} & \textbf{Precision (\%)} & \textbf{Recall (\%)} \\
			\midrule
			Background        & 88.63             & 93.97                  & 94.26                   & 93.69                \\
			Water             & 68.80             & 81.52                  & 80.77                   & 82.28                \\
			\bottomrule
	\end{tabular}}
	\label{tab:source_perf}
\end{table}

As can be seen from Table~\ref{tab:source_perf}, the foundational model A1 achieved good segmentation results on its validation set. For the critical "Water" class, the Intersection over Union (IoU) reached 68.80\%, and the F1-score reached 81.52\% (note: in binary semantic segmentation, the F1-score is equivalent to the Dice coefficient). The precision and recall were 80.77\% and 82.28\%, respectively. These metrics indicate that Model A1 has developed strong generalisation capabilities, laying a solid foundation for the subsequent knowledge transfer to the target domain.

\begin{figure}[htbp]
	\centering
	\includegraphics[width=\textwidth]{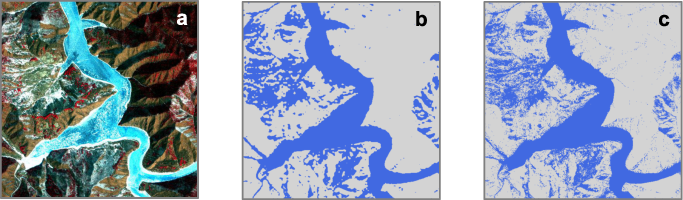} 
	\caption{Segmentation results of the foundational model A1 on a typical scene from the source dataset A: (a) Original Image, (b) Ground Truth, (c) Prediction of Model A1.}\label{fig:source_result}
\end{figure}

Figure~\ref{fig:source_result} displays the segmentation results of the foundational model A1 in a typical scene from the source domain, aiming to visually assess its performance as a starting point for transfer. In this scene's remote sensing image (Fig.~\ref{fig:source_result}a), the water-land boundary is clear, providing a good basis for model learning. By comparing the prediction (Fig.~\ref{fig:source_result}c) with the ground truth (Fig.~\ref{fig:source_result}b), it is evident that the predicted contours highly coincide with the actual water boundaries, demonstrating excellent pixel-level segmentation accuracy. It is noteworthy that the model not only accurately delineated the meandering form of the main river channel but also successfully captured the small tributary flowing into it, indicating a good understanding of the spatial connectivity and structural hierarchy of the water system.

\subsection{Comparative Analysis of Transfer Learning}\label{subsec:transfer_analysis}

To comprehensively evaluate the effectiveness of the proposed transfer learning strategy, a rigorous quantitative and qualitative comparison of models trained under different strategies was conducted on the validation set of the target domain (Dataset B). The strategies include: 1) \textbf{Direct Transfer}, where the foundational model A1 (SegFormer) trained on the source domain was directly applied to the target domain; 2) \textbf{Train from Scratch with SegFormer} (B-scratch (Seg)), using ImageNet pre-trained weights as initialization but trained solely on the target domain; 3) \textbf{Train from Scratch with U-Net} (B-scratch (U-Net)), a simpler baseline model trained from random initialization on the target domain; and 4) \textbf{Transfer Learning} (A2), the proposed strategy of fine-tuning Model A1 on the target domain.

\begin{table}[htbp] 
	\tbl{Performance comparison of the models on the validation set of target dataset B.}
	{\begin{tabular}{@{}l l c c c c@{}}
			\toprule
			\textbf{Model} & \textbf{Class} & \textbf{IoU (\%)} & \textbf{F1-score (\%)} & \textbf{Precision (\%)} & \textbf{Recall (\%)} \\
			\midrule
			A1 (Direct Transfer) & Background & 72.06 & 83.76 & 75.25 & 94.43 \\
			& Water      & 25.50 & 40.64 & 69.24 & 28.76 \\
			\midrule
			B-scratch (SegFormer) & Background & 69.58 & 82.06 & 77.38 & 87.35 \\
			& Water      & 37.47 & 54.51 & 64.40 & 47.26 \\
			\midrule
			B-scratch (U-Net)     & Background & 72.00 & 84.34 & 83.11 & 84.34 \\
			& Water      & 48.82 & 65.61 & 66.65 & 64.61 \\
			\midrule
			\textbf{A2 (Fine-tuned)} & Background & \textbf{77.59} & \textbf{87.38} & \textbf{89.80} & \textbf{85.09} \\
			& Water      & \textbf{64.84} & \textbf{78.67} & \textbf{75.25} & \textbf{82.42} \\
			\bottomrule
	\end{tabular}}
	\label{tab:target_perf}
\end{table}

The updated results, as shown in Table~\ref{tab:target_perf}, reveal performance differences across strategies. The direct transfer model (A1) performed the poorest, with a water IoU of 25.50\% and a recall of 28.76\%. This drop (compared to its 68.80\% IoU in the source domain) highlights the domain shift issue.

The SegFormer model trained from scratch (B-scratch (Seg)) achieved a water IoU of 37.47\%. In comparison, the U-Net model trained from scratch (B-scratch (U-Net)) reached a water IoU of 48.82\%.

The fine-tuned model (A2) achieved the highest performance, with a water IoU of 64.84\%, representing a relative gain of approximately 154\% over direct transfer, 73\% over SegFormer scratch, and 33\% over U-Net scratch.

\begin{figure*}[htbp] 
	\centering
	\includegraphics[width=\textwidth]{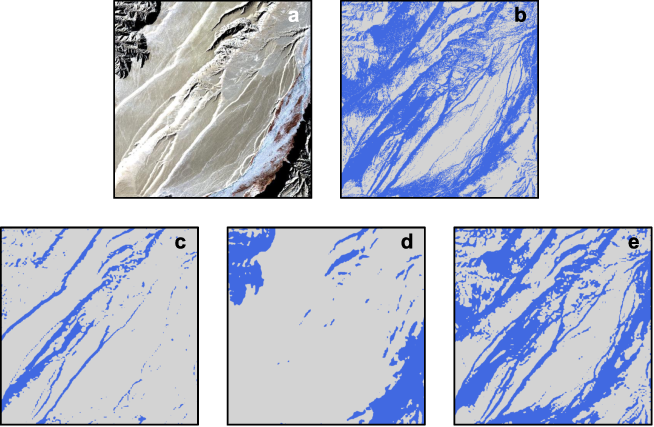} %
	\caption{Segmentation comparison of different models on a typical scene from the target dataset B: (a) Original Image, (b) Ground Truth, (c) Prediction of Model A1 (Direct Transfer), (d) Prediction of Model B-scratch (SegFormer), (e) Prediction of Model A2 (Fine-tuned). The figure focuses on SegFormer-based strategies, with U-Net results provided quantitatively in the table.}\label{fig:target_comparison}
\end{figure*}

Figure~\ref{fig:target_comparison} provides a visual comparison in a challenging Zhada Tulin scene, where water and background spectra are highly similar. The direct transfer (c) shows fragmented errors, SegFormer scratch (d) misses some connectivity, and the fine-tuned A2 (e) restores the river morphology most accurately.

\section{Discussion}\label{sec:discussion}

Our experimental results highlight the critical role of transfer learning in overcoming the dual challenges of domain shift and data scarcity in remote sensing segmentation. The direct transfer model (A1) and the scratch-trained models (B-scratch SegFormer and U-Net) all yielded suboptimal performance, confirming that neither a general model nor a complex architecture alone is sufficient. The proposed two-stage fine-tuning strategy, however, achieved a water IoU of 64.84\%, demonstrating its effectiveness. The visual results in Figure~\ref{fig:target_comparison} corroborate these findings, showing that only the fine-tuned model (A2) could accurately restore the river's morphology against a complex background.

The superior performance of our approach stems from its effective knowledge reconstruction process \citep{pan2010q, yosinski2014transferable}. The source-domain pre-training endowed the model with robust low-level feature extraction capabilities, honed on diverse data that even included significant artifacts like cloud masks. This pre-training made the model resilient to real-world data imperfections. Subsequently, the target-domain fine-tuning allowed the model to adapt its high-level semantic understanding to the specific spectral and geomorphological characteristics of the Zhada Tulin, including high turbidity and the unique sensor properties of GF-2 imagery. This "general-to-specific" adaptation proved crucial for success.

\subsection{Geoscientific Implications of the Segmentation Results}

Beyond the technical advancements, the high-precision water body map generated by our model provides significant insights into the geoscientific characteristics of the Zhada Tulin area \citep{li2013china, zhu2019qing, chen2009china}. The segmentation results reveal that the water systems exhibit a distinct dual control by tectonics and lithology. The main channel of the Xiangquan River follows a NW-SE trend, strictly controlled by regional fault structures, while its tributaries are primarily incised into softer lacustrine sediments, forming deep canyons. This drainage pattern serves as a microcosm of the coupled processes of tectonic uplift and fluvial incision during the Quaternary, reflecting the geomorphological evolution of the Tibetan Plateau.

More importantly, our high-precision extraction reveals the highly concentrated nature of water resources in this arid region. Statistical analysis indicates that over 80\% of the water surface area is confined to less than 20\% of the total river channel length, mainly along the Xiangquan River's main stem and its three primary tributaries. This extremely uneven distribution creates a pronounced 'corridor effect',  where the river valleys form relatively humid oases that support the region's limited vegetation cover and wildlife habitats, while the vast surrounding earth forest plateaus remain largely barren. From a climate resilience and disaster risk management perspective, this finding underscores the high dependency of local human activities, such as seasonal grazing, on these few critical water sources. The concentrated water distribution pattern also has important implications for flood risk assessment and early warning systems, as flash floods during intense precipitation events are likely to be concentrated in these narrow corridor zones. Therefore, establishing a long-term monitoring system based on our remote sensing approach is of great practical significance for adapting to climate change and maintaining ecological balance in the region,while also supporting disaster preparedness efforts in this climate-sensitive area.

\subsection{Limitations and Future Directions}

Despite the promising results, this study has limitations. The water IoU of 64.84\%, while a significant improvement, indicates that challenges remain, particularly in scenes with extreme background interference. This is partly attributable to the limited size of the target domain dataset. Although our results are reasonable for a challenging small-sample plateau setting when compared to benchmarks on other datasets (e.g., ~70\% IoU on LoveDA \citep{wang2021loveda}), further improvements could be achieved with more data.

Our study primarily validated the classic pre-train-fine-tune paradigm. Future research should explore two key directions. First, developing a multi-temporal monitoring system to analyze long-term water dynamics, which would provide deeper insights into the hydrological response to climate change and support enhanced early warning capabilities for climate-related hazards.Second, incorporating more advanced domain adaptation techniques \citep{ganin2015unsupervised, tzeng2017adversarial, long2015learning, ghifary2016deep, saito2018maximum, hoffman2018cycada, pei2018multi}, such as adversarial or self-supervised learning, could further bridge the domain gap with even less labeled data. Extending this framework to other remote sensing tasks, like building or vegetation extraction, also presents valuable opportunities for advancing Earth observation capabilities that support sustainability and climate resilience applications.

\section{Conclusion}\label{sec:conclusion}

This study proposed and validated a two-stage transfer learning strategy based on the SegFormer model to address the critical challenges of domain shift and small-sample learning in remote sensing water body segmentation. By pre-training on a diverse source domain and subsequently fine-tuning on a specific, challenging target domain—the Zhada Tulin—our approach demonstrated significant performance gains. The water body segmentation IoU on the target validation set surged from 25.50\% (for direct transfer) to 64.84\%, substantially outperforming scratch-trained baselines.

Beyond the technical achievement, this research underscores the geoscientific value of high-precision segmentation. The generated water map revealed the highly concentrated spatial distribution of water resources in the Zhada Tulin, a key insight for understanding the hydrological processes and delicate ecological balance in this arid plateau region. This work not only provides a robust and replicable framework for thematic information extraction in data-scarce environments but also offers vital technical support for monitoring the headwaters of transboundary rivers and assessing water security under a changing climate. In conclusion, our study effectively bridges advanced deep learning techniques with pressing geoscientific applications, offering a valuable paradigm for future remote sensing research.

\section*{Author Contributions}
All authors contributed to the study conception and design. Haonan Chen was responsible for the initial investigation, data acquisition, and preprocessing. Building on this foundational work, the formal analysis, methodology development, software implementation, and validation experiments were carried out by Xin Tong. Both authors contributed to the writing, review, and editing of the manuscript and have approved the final version.

\section*{Disclosure of interest}
The authors report no conflict of interest.

\section*{Data availability statement}
The datasets generated and/or analysed during the current study are available from the first author (H.C.) on reasonable request.

\section*{Code availability}
The source code developed for this study is available from the corresponding author (X.T.) on reasonable request.

\bibliographystyle{tfcad}
\bibliography{interactcadsample}

\end{document}